\documentclass[runningheads,a4paper]{llncs}

\usepackage{amssymb}
\usepackage{graphicx}
\usepackage{algorithm}
\usepackage{algorithmic}
\usepackage{hyperref}
\usepackage{caption}
\usepackage{subcaption}
\usepackage[numbers]{natbib}

\usepackage{url}
\urldef{\mailsc}\path|Djallel.Bouneffouf}@it-sudparis.eu|

\begin{document}

\mainmatter  

\title{ Exponentiated Gradient Exploration for Active Learning}

\titlerunning{Lecture Notes in Computer Science. Authors' Instructions}

%
%
\author{Djallel Bouneffouf
}
\authorrunning{Lecture Notes in Computer Science. Authors' Instructions}

\institute{Department of Computer Science, T\'{e}l\'{e}com SudParis, UMR CNRS Samovar, 91011 Evry Cedex, France,\\
\mailsc\\
}

%
%

\toctitle{Lecture Notes in Computer Science}
\tocauthor{Authors' Instructions}
\maketitle

\begin{abstract}
Active learning strategies respond to the costly labelling task in a supervised classification by selecting the most useful unlabelled examples in training a predictive model. 
Many conventional active learning algorithms focus on refining the decision boundary, rather than exploring new regions that can be more informative. 

In this setting, we propose a sequential algorithm named $EG-Active$ that can improve any Active learning algorithm by an optimal random exploration. Experimental results show a statistically significant and appreciable
improvement in the performance of our new approach over the existing active feedback methods.

\end{abstract}

\section{Introduction}
\label{sec.introduction}

In active learning, the learning algorithm has access to a large set of unlabeled examples and some oracle that can query to get the label of an individual example. Queries oracle are assumed to be expensive, it is only feasible to label a small subset. Thus the goal of the algorithm is to actively select examples so that a good hypothesis is learned while using as few labelled examples as possible. 

Many conventional active learning algorithms choose to label points that are near the decision boundary of the current hypothesis. This can function well if the active learner is aware of all the important regions of the instance space, i.e. there are no large examples that the learner's hypothesis will misclassify since it hasn't seen labelled examples from them. Such active learners are good at labelling examples near the boundary to refine it, but they do not conduct a random searching for large regions in the instance space that they would incorrectly classify.

Recent work done \cite{Osugi2005} in this sense considers random exploration in the active learning. The author considers two types of active learner, the first which is dedicated to exploit is based on refining the decision boundary and the second is dedicated to explore (random). More the random exploration gets reward more is the exploration. 

The drawback of this approach is in the fact that, this approach takes a long time to find the optimal random exploration rate. To tackle this problem, in this paper we propose an algorithm named EG-Active which can improve any existing active learning algorithm by using a random exploration which is parametrised with an exponentiated gradient optimization.

We have tested EG-Active with real data where we have observed its performance regarding the state of the art.

The remaining of the paper is organized as follows. Section \ref{sec:related} reviews related works. Section \ref{sec:Active} describes the model and the proposed algorithm. The experimental evaluation is illustrated in Section \ref{sec:experimental}. The last section concludes the paper and points out possible directions for future work. 
  
\section{Related Work}
\label{sec:related}

We refer, in the following to recent works that address Active Learning problem. 

\subsection{Active Learning.} A variety of AL algorithms have been proposed in the literature employing various query strategies. One of the most popular strategy is called the uncertainty sampling (US), where the active learner queries the point whose label is uncertain \cite{Lewis1994}.

The uncertainty in the label is usually calculated using entropy, or variance of the label distribution. The authors in \cite{Seung1992} have introduced the query-bycommittee (QBC) strategy where a committee of potential models, which all agree with the currently labelled data is maintained and, the point where most of the committee members disagree with, is considered for querying. 
Other strategies include the maximum expected reduction in error \cite{Zhu03combiningactive} or variance reducing query strategies such as  \cite{ZhangOles2000} to query the optimal point. 

All the above proposed approaches have just exploited the data and do not consider the random exploration that can help to find the great point to label.

\subsection{Random exploration in Active Learning.} 

Recently, the Random exploration has been used in different domains such as recommender system (RS) and information retrieval. For example, in \cite{BouneffoufBG12, bouneffouf2013drars}, authors model RS as a contextual bandit problem. The authors propose an algorithm which perform random recommendation according to the risk of upsetting the user. However, to our knowledge there has been only one paper addressing the random exploration in active learning. 
Authors in \cite{Osugi2005} address this problem by randomly choosing between exploration and exploitation at each round, and then receive feedback on how effective is the exploration.
The impact of exploration is measured by the induced change in the learned classifier when an exploratory example is labelled and added to the training set. The active learner updates the probability of exploring in subsequent rounds based on the feedback it has received. However none of the optimisation techniques is used to compute the optimal exploration and the work has only been done to improve the uncertainty sampling technique.

\subsection{Our Contributions.} 
As shown above, none of the mentioned works propose to improve any active learning by random exploration. This is precisely what we intend to do by exploiting the following new features: 

(1) We propose a new generic algorithm named EG-Active, that can improve the results of any active learning algorithm by considering the exploration at each iteration. 

(2) We propose to parametrize this exploration using an exponentiated gradient that allows the EG-Active using the optimal exploration.
 
\section{Active Learning with Random Exploration}
\label{sec:Active}
This section focuses on the proposed model, starting by introducing the key notions used in this paper.

\textbf{Pool based AL.} 

In pool based AL we are provided with a pool $X = \{x_1, ... x_n\}$ of unlabelled points and a labelling oracle $O$, which when queried for the label of $x$, returns $ y \sim P_Y|X=x$. 
Algorithms in the pool based setting have to decide which points to query by looking at the entire pool.

\textbf{Reward.} 

A metric is used to measure the variation of the hypothesis learned by the model between two iterations. More the hypothesis learned by the model varies more is the reward.
We now define the function $d(h, h')$ that we use to get the variation of the model.

Let $X = {x_1, . . . , x_m} = L \cup U$ be the set of labelled and unlabelled training examples that we have. Then for each of the two real-valued hypotheses $h(.), h'(.)$, we define the vectors $H = (h(x_1), h(x_2), . . . , h(x_m))$ and $H' = (h'(x_1), h'(x_2), . . . , h'(x_m))$, i.e. vectors of the real-valued predictions of $h$ and $h'$ on X. 

Now we define $d(h, h')$, 
\begin{equation}
\label{eq:d}
d(h, h')=\frac{H.H'}{||H||.||H'||}\\ 
\end{equation} 
In Eq.~\ref{eq:d}, we compute the cosin similarity between the two vectors $H$ and $H'$. Thus $d(h, h') \in [−1, +1]$ is the cosine of the angle between $H$ and $H'$, and we normalise le ratio of classes in the interval [0, 1] using the Eq.~\ref{eq:Rt}.
\begin{equation}
\label{eq:Rt}
r_t=\frac{2.cos^{-1}(d(h|h'))}{\pi}\\ 
\end{equation} 

\subsection{$\epsilon$-Active.} In order to improve the result of any active learning algorithm, we propose to overlap the any existing algorithm by an algorithm that consider at each iteration a random exploration $\epsilon$.  
 
\begin{algorithm}[H]
\caption{$\epsilon$-Active}
\label{alg:rucb} 
\begin{algorithmic}[1]
\STATE {\bfseries Input:} $X, \epsilon$  
\STATE {\bfseries Output:} $x_t$ 
\STATE $  x_t =\left\{ \begin{array}{rcl}
   		Activelearning(X)   & if( q < \epsilon) \\ 
        Random(X)  & if( q \geq \epsilon) 
             \end{array}\right. $ 
\STATE {\bfseries } \textbf{if} $x$ was not queried in the past \textbf{then} Query $O$ for label $y$ of $x$
\STATE {\bfseries } Observe reward $r_t $                  
    
   \end{algorithmic}
\end{algorithm}
In Algorithm~\ref{alg:rucb}, $Activelearning$ can be any existing active learning for example query by committee, uncertainty sampling or others. 

\subsection{Computing the optimal random exploration.} To consider the random exploration on the Active learning algorithms, the proposed method update the exploration value $\epsilon$ dynamically. 

In each iteration, the algorithm runs a sampling procedure to select a new $\epsilon$ from a $\epsilon$ finite set of candidates. The Probabilities associated to the candidates are uniformly initialized and updated with the Exponentiated Gradient (EG) \cite{10}. This updating rule increases the probability of a candidate $\epsilon$ if it leads to a user's click. 

First we assume that we have a finite number of candidate values for $\epsilon$, denoted by ($\epsilon_1$, ...,  $\epsilon_T$ ), and we try to learn the optimal $\epsilon$ from this set. To this end, the $EG$-Active introduce $p = (p_1, ..., p_T $), where $p_i$ stands for the probability of using $\epsilon_i$ in the $\epsilon$-Active algorithm. These probabilities are initialized to be $\frac{1}{T}$ at the beginning and then iteratively updated through iterations. 

The algorithm use a set of weights $w = (w_1, ..., w_T )$ to keep track of the performance of each $\epsilon_i$ and update them using the EG algorithm. The idea is to increase $w_i$ if the algorithm receives a click from using $\epsilon_i$. Finally, the algorithm calculates $p$ by normalizing $w$ with smoothing. Algorithm~\ref{alg:EG-greedy} shows the $EG$-Active.

\begin{algorithm}[H]
\caption{EG-Active}
 \label{alg:EG-greedy}
 \begin{algorithmic} 
  \STATE {\bfseries Input:} $(\epsilon_1, ..., \epsilon_T):$ candidate values for $\epsilon$ \\
  \STATE $\beta$, $\tau$ and $k$: parameters for EG \\ 
  \STATE  $N$: number of iterations
   \STATE $p_k \Leftarrow \frac{1}{T}$ and w$_k \Leftarrow 1$, $k = 1, ... ,T$
   \FOR{i=1 {\bfseries to} N}
   \STATE  Sample $d$ from discrete $(p_1, ... , p_T )$
   \STATE  Run the $\epsilon$-Active with $\epsilon_d$
   \STATE  Receive the feedback $r_t$
   \STATE  $w_k \Leftarrow w_k$ exp($\frac{\tau [r_i I(k = d)+\beta]}{p_k}$ ), $k = 1, ... , T$
   \STATE  $p_k \Leftarrow (1 - k) (\frac{w_k}{\sum_{j=1}^{T}w_j} + \frac{k}{T})$, $k = 1, ... , T$
    \ENDFOR
   \end{algorithmic}
\end{algorithm}
In Algorithm~\ref{alg:EG-greedy}, $I[z]$ is the indicator function and $\tau$ and $\beta$ are smoothing factors in weights updating. $k$ is a regularization factor to handle singular $w_i$. 
\section{Experimental Evaluation}
\label{sec:experimental}
To conduct our evaluation, we have got from our company a corpus containing utterances in French of a typical communication between a customer and a call center of a Telecom company.
 
There are 7 765 utterances annotated by human experts that have been collected in four different datasets. The unannotated part consists of 3 911 695 utterances.

We use a corporate supervised algorithm (rule based algorithm). We simulate in the experiments an expert (oracle) on the unannotated corpus by using the rule based algorithm which is trained by 7 765 uttrances. Note that the objective of this evaluation is to observe the improvement that can add the random exploration to the existing active learning.

In our experiments, we consider a version of the rule based algorithm without training where at each iteration the active learning tries to selects from the unannoteted the most interesting utterances to annotate and integrate in the training set of the rule based algorithm.

By relating the results to the newer versions, one can verify the usefulness of the proposed approach. Moreover, 
We calculate the regret every 100 iterations and we run the process during 2000 iterations which correspond to our budget in term of labelling. 

In addition to the random (baseline), we compare our methods by constructing 3 groups of algorithms: the first group is the state of the art algorithm described in the related work (Sec. \ref{sec:related}) which are the Sampling by committee, request uncertainty sampling and density weight methods. 

The second group contains a modified state of the art algorithms, where we have added a fixed random exploration to the existing state of the art algorithms, for example 0.5-QBC means that with probability 0.5 the algorithm does random exploration and otherwise does request by committee. 

The third group contains the proposed model EG-Active tested with different existing algorithm, for example EG-Active(committee) means that we have used the Sampling by committee in our model.

In the fourth group we have added a dynamic random exploration to the existing state of the art algorithms as it is done in \cite{Osugi2005}, for example P-QBC is an algorithm that uses \cite{Osugi2005} strategy to compute the probability of the random exploration.

\begin{figure} [h!]       
        \begin{subfigure}[b]{0.8\textwidth}
                \includegraphics[width=1.3\textwidth]{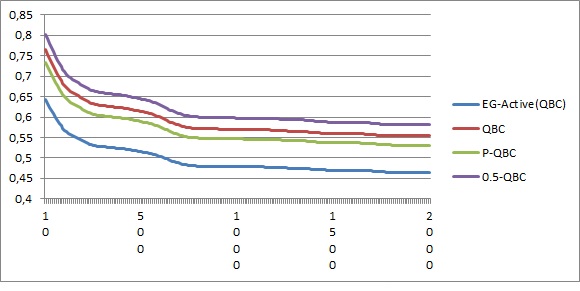} 
                \label{fig:gull}
        \end{subfigure}%
        ~ 
          
        \begin{subfigure}[b]{0.8\textwidth}
                \includegraphics[width=1.3\textwidth]{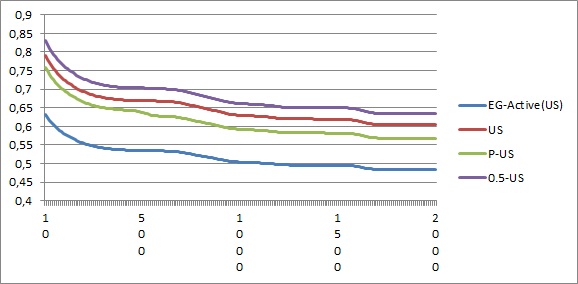} 
                \label{fig:tiger}
        \end{subfigure}
                \begin{subfigure}[b]{0.8\textwidth}
                \includegraphics[width=1.3\textwidth]{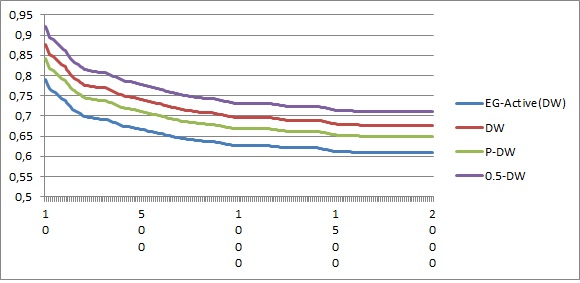} 
                \label{fig:tiger}
        \end{subfigure}
        \caption{Average Regret for Active learning algorithms}\label{fig.average}
\end{figure}

In Fig. ~\ref{fig.average}, the horizontal axis represents the number of iterations and the vertical axis is the performance metric.

We have several observations regarding the different Active learning algorithms.
We observe from the plot that a fixed and not tuned random exploration lead to a bad result. This confirms that a pure exploration is not interesting, and it confirms the need for a dynamic random exploration tuning.
 
A dynamics exploration leads to an improvement result of the active learning as it is shown by P-US, P-CBQ and P-WD. 
As expected, EG-Active(US), EG-Active(CBQ) and EG-Active(WD) effectively have the best convergence rates.

EG-Active(US), EG-Active(CBQ) and EG-Active(WD) decrease the average regret respectively by a factor of 0,84, 1,3, 0,9 over the baseline. The improvement comes from an optimization strategy for defining exploration. 

These algorithms rapidly find the optimal random exploration to use, which is not the case of the P-US, P-CBQ and P-WD that takes more time.  
 
\section{Conclusion}
\label{sec:length}
In this paper, we have proposed an improvement of the active learning by considering a random exploration.  
We have validated our work with data from real-world application and shown that the proposed model offers promising results. 
This study yields to the conclusion that considering the great chosen random exploration used with any active learning algorithm increases its result. In considering these results, we plan to investigate public benchmarks, we also plan to study this random exploration in the bandit algorithms.

\bibliographystyle{abbrv} 
\bibliography{typeinst} 

\end{document}